\documentclass[runningheads]{llncs}
\usepackage{graphicx}
\usepackage{hyperref}
\usepackage{microtype}
\usepackage{booktabs}
\usepackage{xcolor}
\begin{document}
\title{Domain Randomization for Object Counting\thanks{This project has received funding from the European Union’s Horizon 2020 research and innovation programme under the Marie Skłodowska-Curie grant agreement No 765140. This publication has emanated from research supported by Science Foundation Ireland (SFI) under Grant Number SFI/12/RC/2289\_P2, co-funded by the European Regional Development Fund.}}
%
%
\author{Enric Moreu\inst{1,2}\orcidID{0000-0002-0555-3013} \and
Kevin McGuinness\inst{1,2}\orcidID{0000-0003-1336-6477} \and
Diego Ortego\inst{1,2}\orcidID{0000-0002-1011-3610}\and
Noel E. O'Connor\inst{1,2}\orcidID{0000-0002-4033-9135}}
\authorrunning{Enric Moreu et al.}
%
\institute{Insight Centre for Data Analytics, Ireland \and
Dublin City University, Ireland
}
\maketitle              
\begin{abstract}
Recently, the use of synthetic datasets based on game engines has been shown to improve the performance of several tasks in computer vision. However, these datasets are typically only appropriate for the specific domains depicted in computer games, such as urban scenes involving vehicles and people.
In this paper, we present an approach to generate synthetic datasets for object counting for any domain without the need for  photo-realistic techniques manually generated  by expensive teams of 3D artists. 
We introduce a domain randomization approach for object counting based on synthetic datasets that are quick and inexpensive to generate. We deliberately avoid photorealism and drastically increase the variability of the dataset, producing images with random textures and 3D transformations, which improves generalization. Experiments show that our method facilitates good performance on various real word object counting datasets for multiple domains: people, vehicles, penguins, and fruit. The source code is available at:  \url{ https://github.com/enric1994/dr4oc}

\end{abstract}

\section{Introduction}
Object counting is a  computer vision task, the goal of which is to automatically estimate the number of objects in an image or video. It has gained a lot of interest in  recent years because of its many potential uses: it can help to identify the congestion level in a shopping center~\cite{zhang2016single} (people counting), the level of traffic on a road~\cite{TRANCOSdataset_IbPRIA2015} (vehicle counting),  the status of a penguin colony~\cite{arteta2016counting} (habitat monitoring), or even to monitor a harvest~\cite{rahnemoonfar2017deep} (fruit counting).

The  main challenge of object counting is that the model has to learn all the variations of the objects in terms of their size, shape, and pose whilst also dealing with occlusion and perspective effects.
Furthermore, object counting algorithms tend to overfit because of the small amount of annotated data available, which degrades their performance when applying the model on other slightly different domains. 

To address the problems above, some computer vision algorithms are trained or pretrained using synthetic data, which can be automatically annotated with perfect precision for a range of application domains, including those where collecting data  is problematic. Many computer vision tasks such as optical flow, detection, segmentation, or counting have benefited from the use of synthetic data. There are many well-known artificially generated datasets \cite{richter2016playing} that are particularly useful because of their size, quality of the annotations, and the variability within the dataset. 

The task of developing novel approaches to people counting in particular has benefited from the use  of synthetic data. However, when counting  objects from other domains such as wildlife, food, or everyday arbitrary objects, the datasets produced by game engines are not useful because there are no realistic video game renderings of these types of objects. It is not practical to create realistic datasets for many different tasks because of the significant manual effort and production costs required~\cite{radau2009evaluation}.

Furthermore, models trained with synthetic images from a particular domain perform poorly when tested on a different target domain because of the domain gap, which has posed considerable obstacles to real-world adoption of synthetic data for computer vision applications. The main cause is that convolutional neural networks (CNN) introduce a bias towards textures, memorizing them instead of shapes~\cite{geirhos2018imagenet}. For object counting, understanding the shape of the items is of paramount importance in order to address the challenges of overlapping objects and occlusions. 

\begin{figure}[t]

\begin{minipage}[b]{0.48\linewidth}
  \centering
  \centerline{\includegraphics[width=4.0cm]{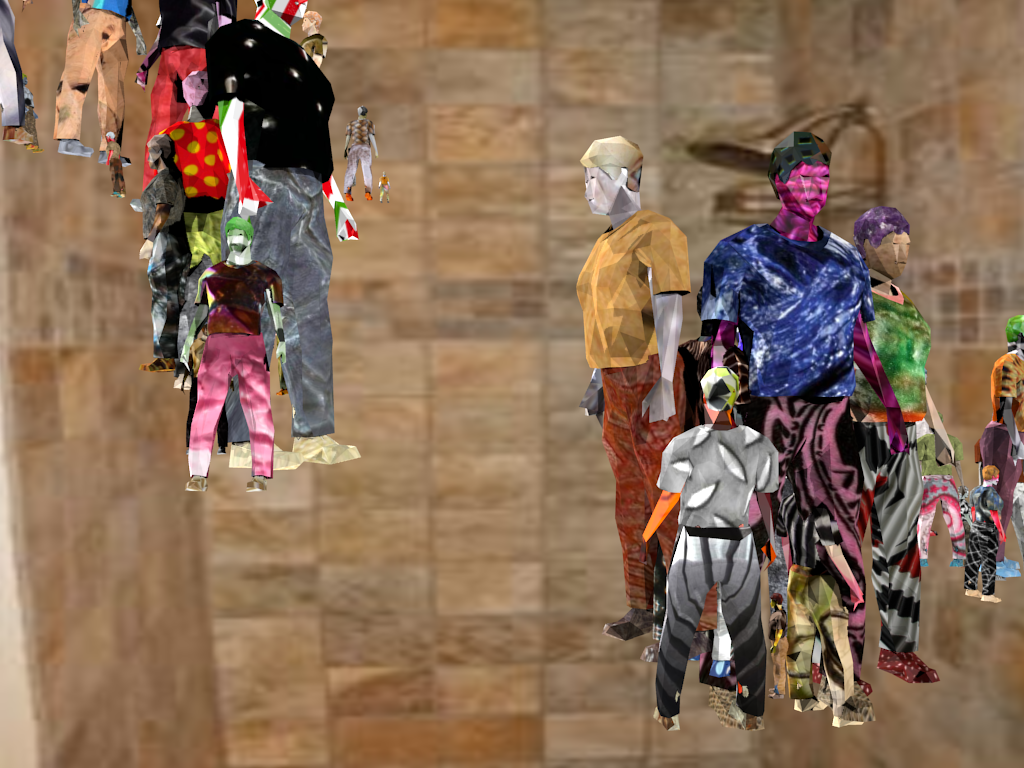}}
  \centerline{(a) Crowd counting}\medskip
\end{minipage}
\hfill
\begin{minipage}[b]{0.48\linewidth}
  \centering
  \centerline{\includegraphics[width=4.0cm]{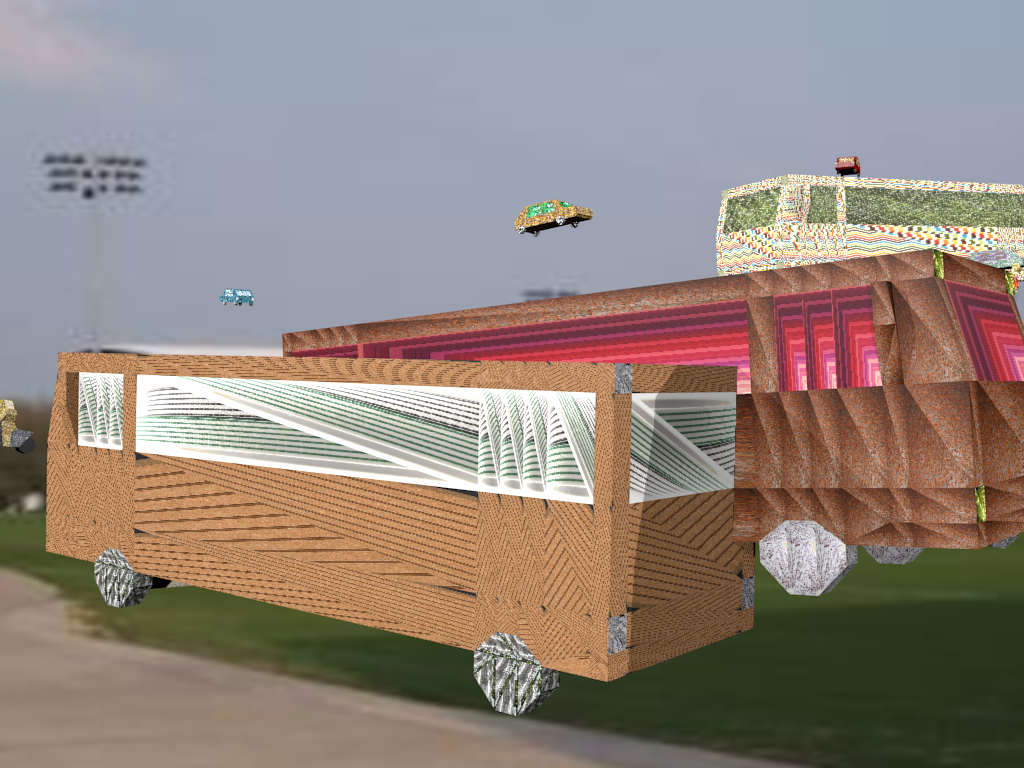}}
  \centerline{(b) Vehicle counting}\medskip
\end{minipage}
\hfill
\begin{minipage}[b]{0.48\linewidth}
  \centering
  \centerline{\includegraphics[width=4.0cm]{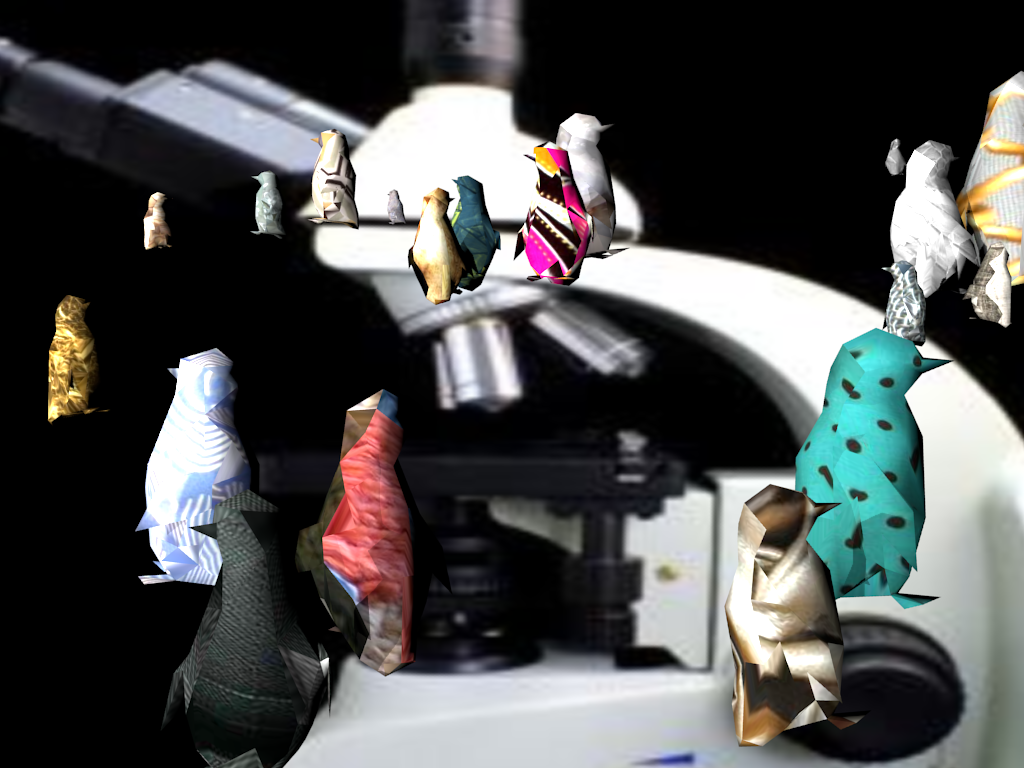}}
  \centerline{(c) Environmental survey}\medskip
\end{minipage}
\hfill
\begin{minipage}[b]{0.48\linewidth}
  \centering
  \centerline{\includegraphics[width=4.0cm]{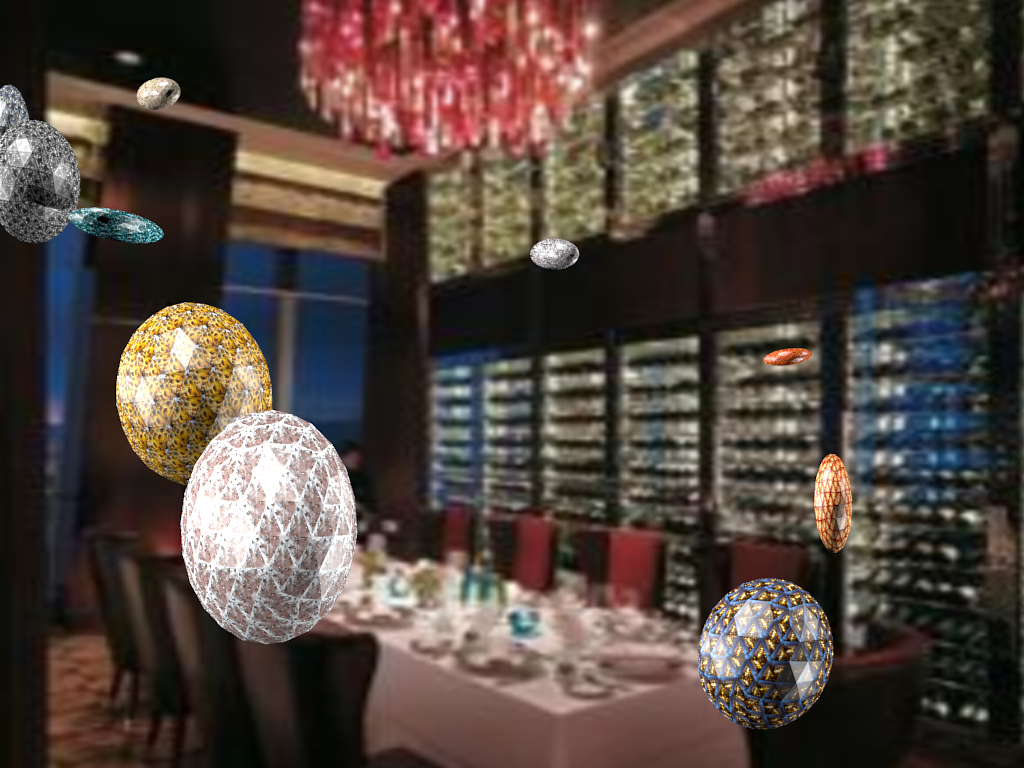}}
  \centerline{(d) Harvest study}\medskip
\end{minipage}
\caption{Synthetic images generated using the proposed domain randomization approach.}
\label{fig:samples}
\end{figure}

Domain randomization (DR)~\cite{tobin2017domain} can reduce the impact of the domain gap  by generating highly variable samples at the cost of increasing the complexity of the task. The objective of increasing variability is to expand the spectrum of possibilities of the source domain whereby the real-world domain becomes just another variation. The synthetic samples generated with this technique tend to look less realistic because of the random textures, lighting, and backgrounds used. DR avoids photorealism, minimizing the need for artistic design. Figure~\ref{fig:samples} shows several examples of DR applied to different environments. 

The contributions of this paper are as follows:
\begin{itemize}

\item
We train an object counting algorithm without labeling any data. The ground truth is calculated automatically during the generation of the synthetic images. 

\item We introduce the first domain randomization approach for object counting based entirely on synthetic images. We increase the variability of the synthetic dataset by applying random textures, backgrounds, and lighting effects to the 3D scene. We demonstrate good performance on real-world datasets that is consistent across multiple domains.

\item We introduce a set of 3D transformations that increase the variability of the 3D models while preserving their inner shape, making the task more complex during training but improving  generalization at test time. To the best of our knowledge, we are the first to use 3D transformations to randomize synthetic images in this way.

\end{itemize}

\section{Related work}

Early object counting algorithms  mainly targeted crowd counting. They applied detection-based approaches such as R-CNN~\cite{he2017mask} and YOLO~\cite{redmon2016you} to estimate the number of people in an image  and demonstrated reasonable accuracy in sparse scenes~\cite{ren2017novel}. However, the performance dropped on densely crowded scenes where people overlapped with heavy occlusion. Alternative  regression-based methods~\cite{idrees2013multi} can extract features (textures, gradients, shapes) to overcome occlusion and learn a mapping function to evaluate how sparse the scene is, but they ignore the spatial information. In  general, CNN-based approaches~\cite{pham2015count} predict density maps to estimate the number of instances in the scene and use the spatial information contained in the density map. Currently, most of the object counting state-of-the-art algorithms are based on fully convolutional networks~\cite{zhang2018crowd} combined with other techniques such as analyzing the context~\cite{guo2019dadnet}, using the perspective information~\cite{gao2019pcc} or applying a multi-column architecture~\cite{zhang2016single}.

Many object counting datasets have appeared in recent years~\cite{zhang2016single,arteta2016counting,TRANCOSdataset_IbPRIA2015,hani2020minneapple}, especially for crowd counting. In general they are annotated with dots indicating the position of the objects, e.g.~crowd counting datasets define a dot on the head of each person. The annotation of object counting datasets is expensive because it requires precise dot annotations performed by an expert; hence the datasets tend to be small, as shown in Table~\ref{fig:datasets}.

Performance is measured using two main metrics: MAE (mean absolute error) and MSE (mean squared error). They compute the average $L^1$ and $L^2$ distance between the predicted count and the ground truth respectively. MAE and MSE are scale-dependent and therefore can not be used to make comparisons between datasets using different scales, e.g.~it can not be used to compare performance between different counting datasets because they may have a different average of objects per image. The formula used to compute them is described as follows:
\begin{equation}
MAE=\sum_{i=1}^{n} \frac {|{\mathbf x _{i}} - {\mathbf y _{i}}|} n,
\end{equation}
\begin{equation}
MSE=\sum_{i=1}^{n} \frac {({{\mathbf x _{i}} - {\mathbf y _{i}})}^2} n,
\end{equation}
where $\mathbf x$ is the predicted count, $\mathbf y$ is the ground truth and $n$ is the number of images evaluated.

\begin{table}[t]
\centering

\caption{Details of the real-world datasets used to evaluate the proposed object counting method.}

\begin{tabular}{llll}
\toprule
Dataset & Images & Resolution & Avg. Count \\
\midrule
SHT A~\cite{zhang2016single} &      482     &       589 $\times$ 868    &   501.4   \\ 
SHT B~\cite{zhang2016single}  &       716     &      768 $\times$ 1024  &   123.6  \\ 
Penguins~\cite{arteta2016counting}  &      80095      &      2048 $\times$ 1536   &       7.2 \\ 
TRANCOS~\cite{TRANCOSdataset_IbPRIA2015}     &     1244       &        640 $\times$ 480    &    36.5      \\ 
MinneApple~\cite{hani2020minneapple}  &   670   &     720 $\times$ 1280     &     42.1     \\ 
\bottomrule
\end{tabular}
\label{fig:datasets}
\end{table}

Recently, synthetic datasets \cite{richter2016playing} have been used to train deep networks for computer vision. The environments used to create the synthetic datasets range from very simple methods using basic shapes and colors \cite{rahnemoonfar2017deep}, to scenes generated by complex game engines~\cite{richter2016playing} that render photo-realistic images and videos. The main benefit is that the data is labeled automatically, saving a substantial amount of time especially in densely annotated datasets such as those for segmentation~\cite{Cordts_2016_CVPR}. 
Also, it is possible in a virtual environment to reproduce rare scenes that are hard to capture from the real world e.g.~remote hard to access locations or unusual weather phenomena.

DR aims to make CNNs robust against challenges  posed by novel domains outside the training set, a phenomenon known as the domain gap. It was first explained by Tobin et al.~\cite{tobin2017domain} by varying the texture of the objects, background image, and lighting in a semantic segmentation task. The objective is to generate enough variations of synthetic data that the model views real data as just another variation, even if the variations used for training appear unrealistic to humans. Expanding the spectrum of possibilities also raises the complexity of the task, requiring a model with a higher capacity. If the model is trained on a sufficient number of environments it will interpolate well to novel ones. This method can be considered to be an evolved form of data augmentation.

Conversely, domain adaptation (DA)~\cite{wang2018deep} tries to bring the training data closer to the real-world data~\cite{wang2019learning} by matching the distributions of both datasets and learning the shared properties. When DA is applied on synthetic images, they will look more realistic. DA is useful when data can be easily obtained  by modeling the distribution of the synthetic features to match the real ones. However, many domains cannot benefit from this technique because of the high variability of the data and the low amount of real images available, e.g.~face recognition datasets have a limited amount of infants smiling~\cite{xia2017detecting}. 

\section{Proposed Approach}

\subsection{Scene creation}
\begin{figure}[tb]

\begin{minipage}[b]{0.20\linewidth}
  \centering
  \centerline{\includegraphics[width=2.0cm]{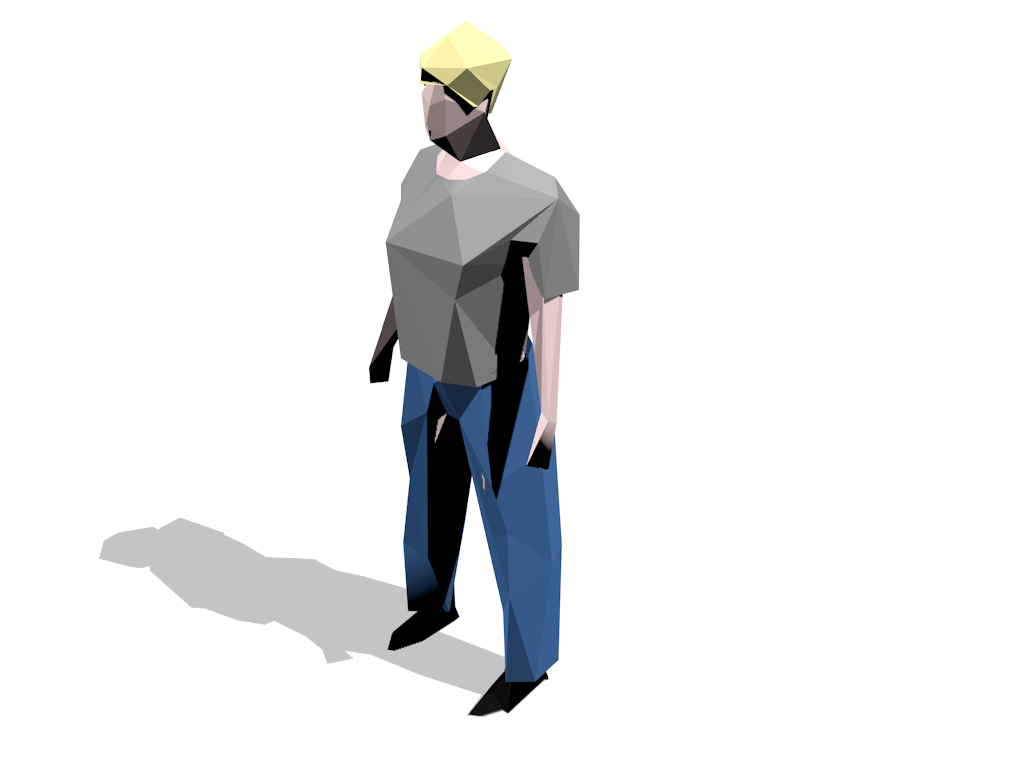}}
  \centerline{(a) Human}\medskip
\end{minipage}
\hfill
\begin{minipage}[b]{0.20\linewidth}
  \centering
  \centerline{\includegraphics[width=2.0cm]{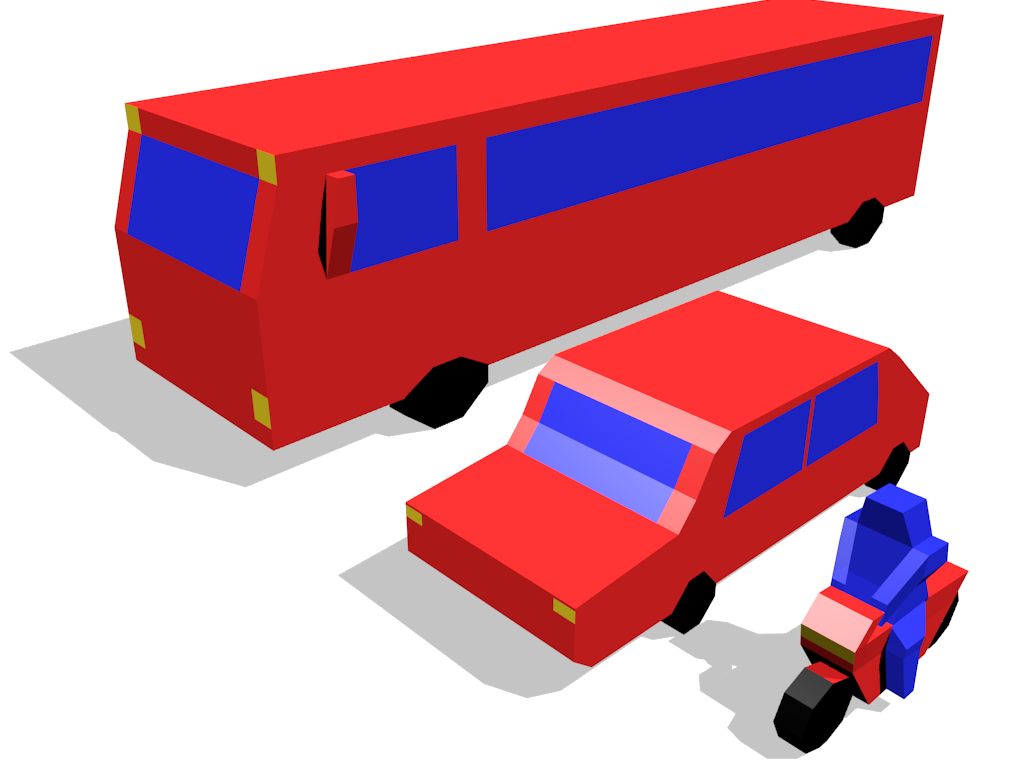}}
  \centerline{(b) Vehicles}\medskip
\end{minipage}
\hfill
\begin{minipage}[b]{0.20\linewidth}
  \centering
  \centerline{\includegraphics[width=2.0cm]{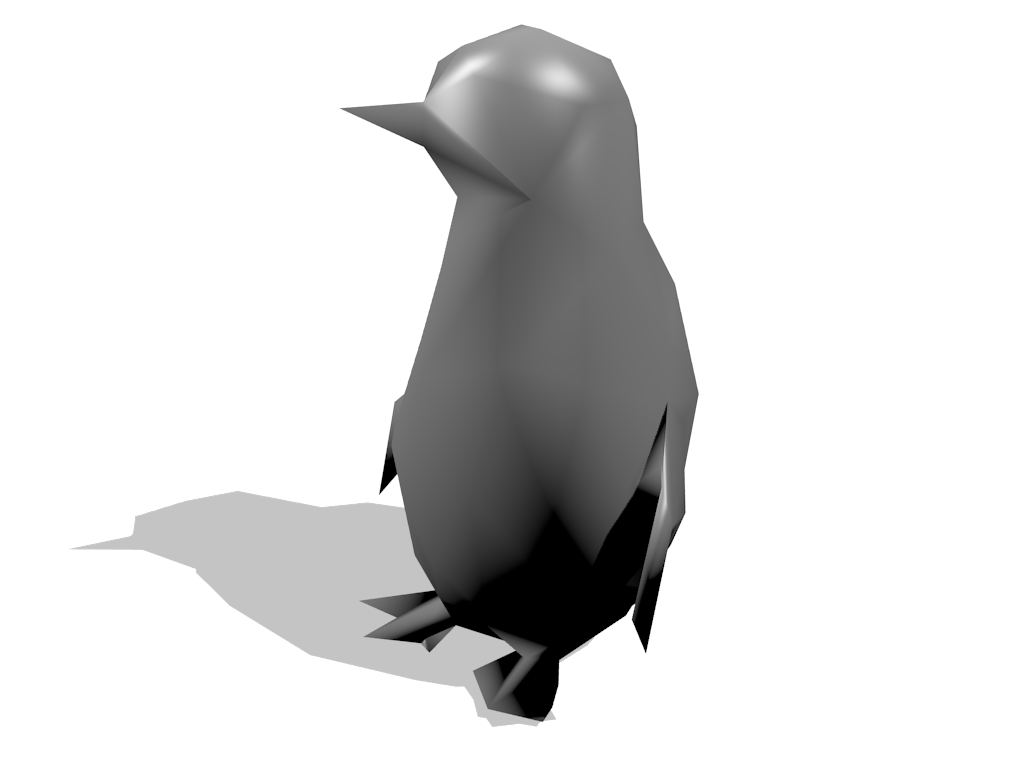}}
  \centerline{(c) Penguin}\medskip
\end{minipage}
\hfill
\begin{minipage}[b]{0.20\linewidth}
  \centering
  \centerline{\includegraphics[width=2.0cm]{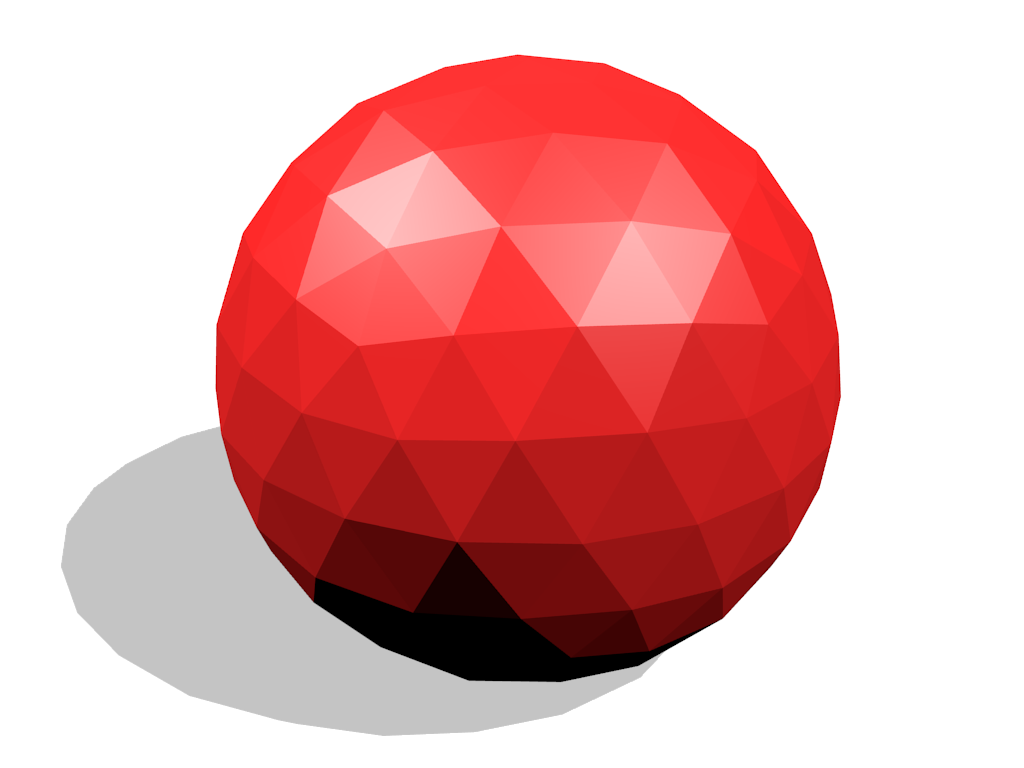}}
  \centerline{(d) Apple}\medskip
\end{minipage}
\caption{Low-poly 3D models used to create synthetic datasets.}
\label{fig:models}
\end{figure}

Our DR datasets are generated using a mixture of 3D models, textures, background images, and lighting effects. The 3D software to render the scenes is Blender~\cite{blender}, which can be easily automated. We specify 3D models to have less than 200 faces, low-poly models, as shown in Figure~\ref{fig:models}. We found that using highly-realistic models, which can have a thousand faces, does not improve the results while significantly increasing the rendering time. 

\begin{figure}[htb]

\begin{minipage}[b]{0.3\linewidth}
  \centering
  \centerline{\includegraphics[width=3.0cm]{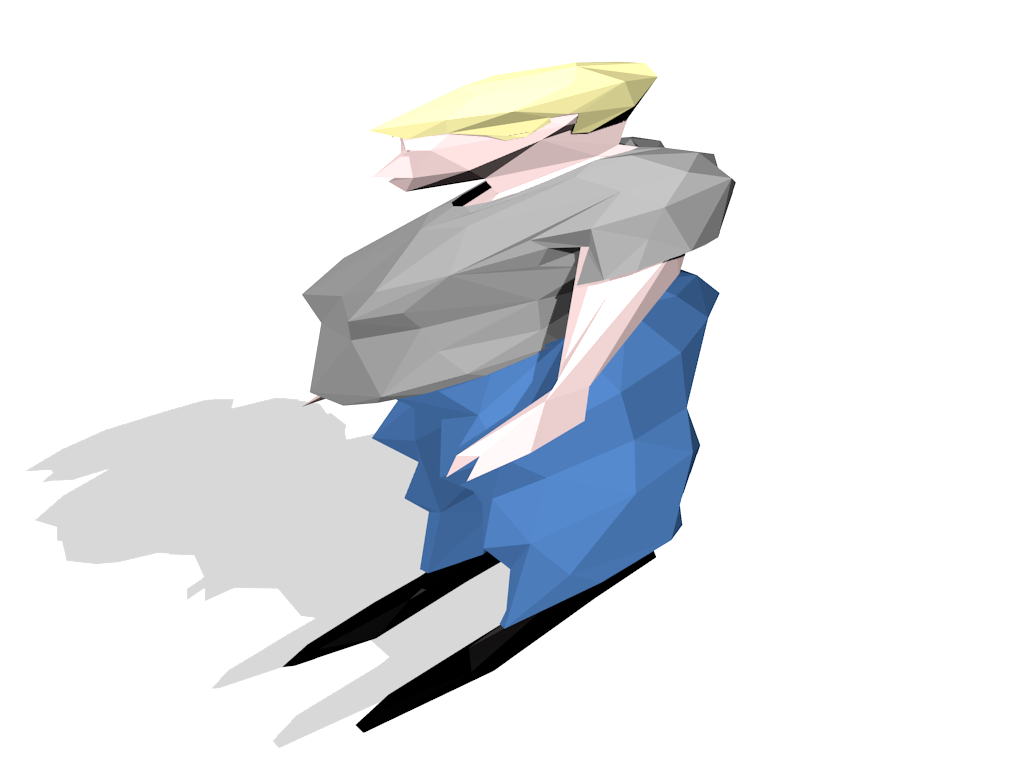}}
  \centerline{(a) Scale}\medskip
\end{minipage}
\hfill
\begin{minipage}[b]{0.3\linewidth}
  \centering
  \centerline{\includegraphics[width=3.0cm]{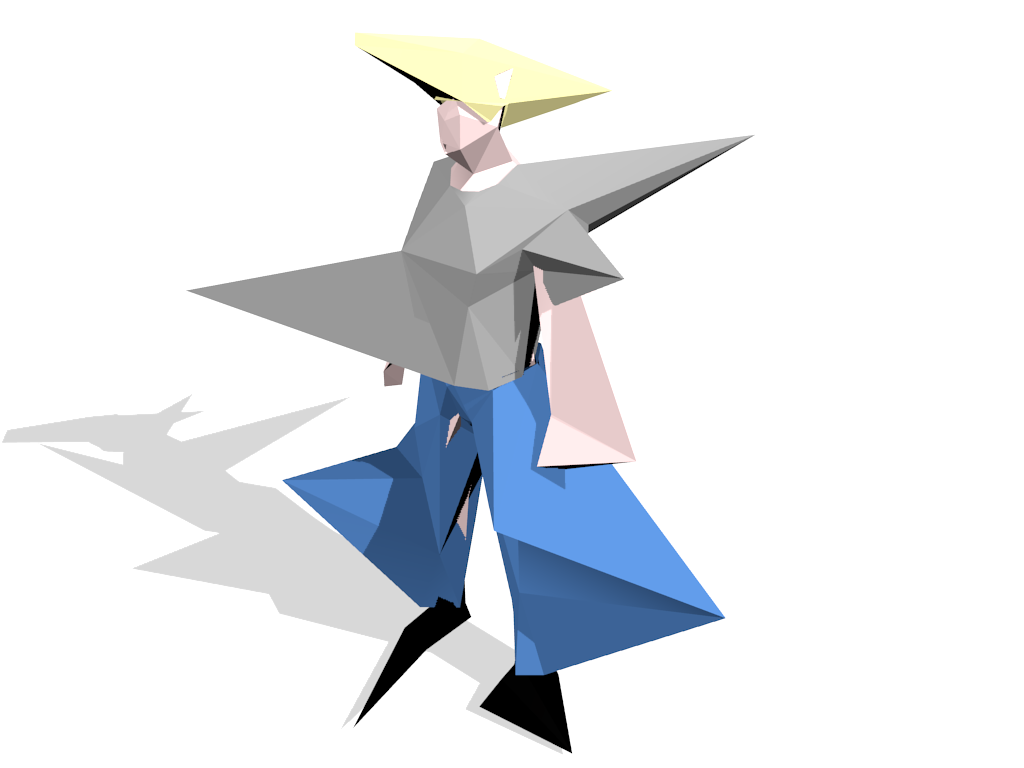}}
  \centerline{(b) Randomization}\medskip
\end{minipage}
\hfill
\begin{minipage}[b]{0.3\linewidth}
  \centering
  \centerline{\includegraphics[width=3.0cm]{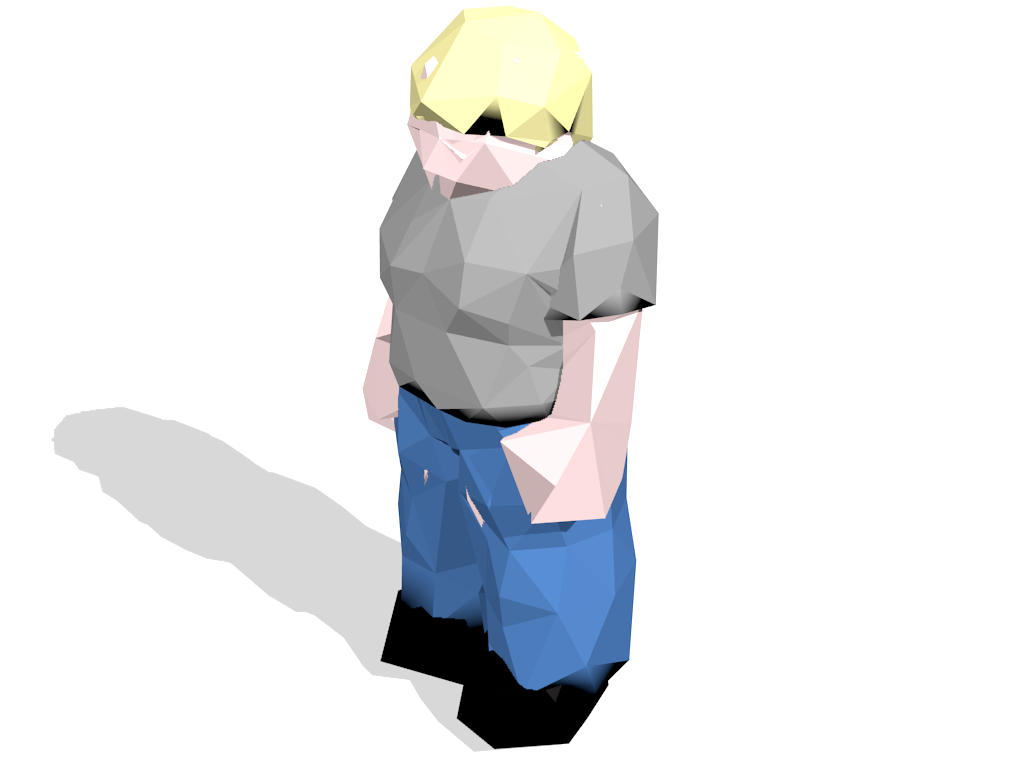}}
  \centerline{(c) Extrude}\medskip
\end{minipage}
\hfill
\caption{3D transformations used for domain randomization. }
\label{fig:transforms}
\end{figure}

The low-poly 3D models structure is modified to produce more variations, some of them unrealistic.
The generated structures are, however, constrained to keep the basic shape, e.g.~humans with one head and two legs, otherwise the model will not learn the inner properties of the object. Learning a vast amount of shapes improves generalization to novel scenarios.
Figure~\ref{fig:transforms} shows the different 3D transformations that we  used to produce the synthetic datasets: scale, randomization, and extrusion. 
Scale smoothly expands/contracts all the vertices on the same axis. It is useful when objects tend to have very different sizes, e.g.~adults tend to be twice as big as children. 

The scale of every 3D model is determined by $K \sim \mathcal U (1/N_o,8/N_o)$ where $\mathcal U$ is a uniform distribution and $N_o$ is the number of objects in the image. Randomizing the vertices of the mesh translates all the vertices in different lengths and directions, uniformly by a factor of 40\%. This method improves the performance in environments where the pose of the objects is variable, e.g.~people can have multiple poses while vehicles do  not.
Extrusion alters the surfaces of the mesh to increase the thickness by adding depth. This helps to make objects bigger or smaller but adds bumps and holes. We used the built-in Solidify transform in Blender and modified the thickness by $T$ where  $T \sim \mathcal U (-0.1,0.5)$.

\begin{figure}[tb]
  \centering
  \centerline{\includegraphics[width=8cm]{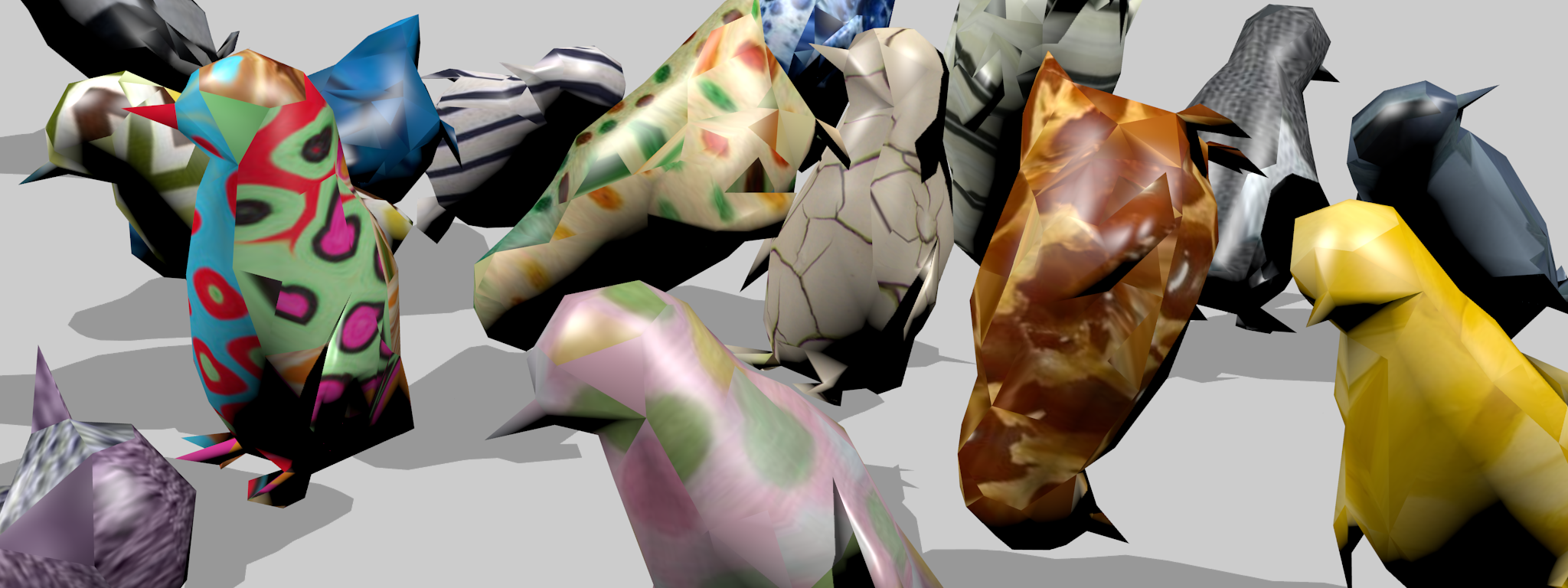}}
  \caption{Random textures from the Describable Textures Dataset applied on a low-poly 3D model of a penguin.}
  \label{fig:textures}
\end{figure}

Textures from the Describable Textures Dataset~\cite{cimpoi14describing} are applied to the 3D models as shown in Figure \ref{fig:textures}. The dataset contains 5640 textures organized into 47 categories. Textures are mapped to the different parts of the 3D models, e.g.~hair, skin, shirt, pants. 
This technique helps our DR approach to transcend realism by producing unrealistic sets of randomly textured 3D models.

The 3D models are placed in the scene by sampling positions from a standard Gaussian mixture  distribution as follows:
\begin{equation}
p(\mathbf x)=\sum_{i=1}^{K}\lambda_i\, \mathcal N({\mathbf x}\mid {\mu _{i}},\Sigma_i),
\end{equation}
 where ${\mathbf x}$ is the three-dimensional $x,y,z$ position, $\lambda_i$ are the mixture component weights, ${\mu_{i}}$ are the means, and $\Sigma_i = I$. The number of components $K$ is sampled for each scene as 
$K \sim \mathcal U (1 + N_o / 20, 2 + N_o/8)$ where $\mathcal U$ is a uniform distribution and $N_o$ is the number of objects in the scene. The mean vectors are uniformly sampled from the rendered area in the 3D space.

This method creates occlusion in the clusters but also produces large empty areas where the background image is displayed. It also mimics how objects are distributed on the real world, e.g.~people are not uniformly distributed \cite{liu2016typical}, they tend to form small groups on the street.
We found that when the objects are distributed uniformly the test mean absolute error (MAE) increased to 63.4 on the SHT B dataset for crowd counting, compared with 23.2 using the Gaussian mixture approach.

Images from the Places2 dataset~\cite{zhou2017places} are used as the background image. The dataset contains a wide range of scenes from 365 different environments including indoors, streets, and nature. The fact that the background images are very different make the task more complex but improves generalization.
Depending on the task, some image categories have been removed to avoid unlabeled instances of the relevant objects in the background, e.g.~the ``stadium-football'' category when counting people or the ``iceberg'' category when counting penguins. Finally, a combination of colored lights is randomly placed around the scene to produce different exposure levels and cast shadows around the 3D objects. 

\subsection{Counting procedure}

We used the Distribution Matching for Crowd Counting~\cite{wang2020DMCount} approach as a baseline. The authors use Optimal Transport to measure the similarity between the normalized predicted density map and the normalized ground truth density map. They also include a total variation loss to force the neighbouring pixels to have similar values. The baseline performance is particularly good on scenes where the density and overlapping of the objects is high. 
We used a ResNet50~\cite{he2016deep} model, pretrained on ImageNet LSVRC~\cite{russakovsky2015imagenet}, as the base model for all of the counting tasks. Horizontal flips are applied to double the amount of available synthetic images. In addition, training images are randomly cropped into multiple smaller images (512 $\times$ 512) to obtain more samples.

\section{Evaluation}

In this section we analyze the performance of our DR method and evaluate the models trained entirely with synthetic data on real-world datasets. Two types of experiments are conducted: 1) testing on real-world datasets; 2) analyzing the effect of 3D transformations.

\subsection{Comparison with real-world datasets }
We propose a new scheme to remedy the lack of datasets in non-urban environments. By training our model only on synthetic data we obtain good performance on multiple real-world environments.
Table~\ref{fig:counting} compares the performance of training with real data and synthetic data. 

\begin{table}
\centering
\caption{Object counting performance (MAE) of real and DR synthetic data on multiple real-world domains.}
\begin{tabular}{lll}
\toprule
Dataset & Real & DR (Ours) \\ \midrule
SHT A~\cite{zhang2016single} &      PGCNet~\cite{yan2019perspective}: 57.0     &       158.7    \\ 
SHT B~\cite{zhang2016single}  &       SANet~\cite{cao2018scale}: 6.5     &      23.2    \\ 
Penguins~\cite{arteta2016counting}  &      Marsden et al.~\cite{marsden2018people}: 5.8      &      14.6         \\ 
TRANCOS~\cite{TRANCOSdataset_IbPRIA2015}     &       FCN-rLSTM~\cite{zhang2017fcn}: 4.2     &        13.6       \\ \bottomrule
\end{tabular}
\label{fig:counting}
\end{table}

Table~\ref{fig:crowd-synthetic} compares our DR approach with Wang et al.~\cite{wang2019learning} for crowd counting. Their method is based on DA applied on images from a realistic video game. Using real-world images to feed a GAN they improve the textures of the video game images.
DA is successfully applied to domains where it is easy to obtain real-world images and produce synthetic data using a video game, e.g.~urban environments involving people and vehicles. Our DR approach obtains similar results without using real-world data and very simple rendering techniques. Whilst the approach of~\cite{wang2019learning} performs better than our approach, it should be noted that our performance is achieved with an automatically generated synthetic dataset.

\begin{table}
\centering
\caption{Crowd counting performance (MAE/MSE) using synthetic data comparison. }
\begin{tabular}{lll}
\toprule
Dataset & GCC~\cite{wang2019learning} & DR (Ours) \\ \midrule
SHT A~\cite{zhang2016single} &      123.4/192.4     &       158.7/253.2    \\ 
SHT B~\cite{zhang2016single}  &       19.9/28.3     &      23.2/41.5    \\ \bottomrule
\end{tabular}
\label{fig:crowd-synthetic}
\end{table}

\subsection{3D transformations analysis}
3D transformations increase the variability of the dataset in terms of shape, improving the generalization to novel domains. To the best of our knowledge, we are the first to use 3D transformations on synthetic datasets for generating training data. Table~\ref{fig:3d-trans} shows how 3D transformations affect performance on the object counting task. For each experiment we generated 2k synthetic images with the given 3D transformation.

We also observed that when applying strong 3D transformations the training process takes longer because the task becomes more complex.

\begin{table}
\centering
\caption{Domain randomization techniques (MAE/MSE).}

\begin{tabular}{llll}
\toprule
Dataset & Scale  & Extrude  & Randomize \\ \midrule
SHT A~\cite{zhang2016single}  &    183.97/291.7   &     159.0/253.2   &  \textbf{158.7/253.2}    \\ 
SHT B~\cite{zhang2016single}  &    29.17/47.8   &     30.2/54.2   &  \textbf{23.2/41.5}     \\ 
Penguins~\cite{arteta2016counting}  &    18.9/25.1   &     \textbf{14.6/20.1}   & 14.6/20.4          \\ 
TRANCOS~\cite{TRANCOSdataset_IbPRIA2015}      &    15.4/19.3   &     \textbf{13.6/17.3}   &  13.6/17.6        \\ 
MinneApple~\cite{hani2020minneapple}    &    22.4/29.4   &     21.3/27.5   &  \textbf{17.5/22.3}        \\ \bottomrule
\end{tabular}
\label{fig:3d-trans}
\end{table}

Randomizing the vertices works better on environments with objects that can present different poses, e.g.~people. The results obtained with the extrude transform are similar to the randomize ones because it also creates small irregularities in the shape. Extrude exhibits good performance on environments where the objects are solid, e.g.~vehicles.

\section{Conclusion}
In this paper, we proposed a domain randomization approach for object counting that can be easily applied to any domain.
The counting model was trained only with synthetic images and achieves good performance on different real-world counting datasets: crowd counting, vehicle counting, penguin counting, and fruit counting. Applying the right 3D transformations to the meshes increases the counting accuracy when evaluating on real-world datasets. The  impact of 3D transformations  depends on the nature of the object, e.g.~variable pose and size.
Future work in this area will look to extend the proposed domain randomization approach to the video domain and to use the depth information from the synthetic data.

%
%
%
\bibliographystyle{splncs04}
\bibliography{citations}

\begin{thebibliography}{10}
\providecommand{\url}[1]{\texttt{#1}}
\providecommand{\urlprefix}{URL }
\providecommand{\doi}[1]{https://doi.org/#1}

\bibitem{arteta2016counting}
Arteta, C., Lempitsky, V., Zisserman, A.: Counting in the wild. In: European
  conference on computer vision. pp. 483--498. Springer (2016)

\bibitem{cao2018scale}
Cao, X., Wang, Z., Zhao, Y., Su, F.: Scale aggregation network for accurate and
  efficient crowd counting. In: Proceedings of the European Conference on
  Computer Vision (ECCV). pp. 734--750 (2018)

\bibitem{cimpoi14describing}
Cimpoi, M., Maji, S., Kokkinos, I., Mohamed, S., , Vedaldi, A.: Describing
  textures in the wild. In: Proceedings of the {IEEE} Conf. on Computer Vision
  and Pattern Recognition ({CVPR}) (2014)

\bibitem{blender}
Community, B.O.: Blender - a 3D modelling and rendering package. Blender
  Foundation, Stichting Blender Foundation, Amsterdam (2018),
  \url{http://www.blender.org}

\bibitem{Cordts_2016_CVPR}
Cordts, M., Omran, M., Ramos, S., Rehfeld, T., Enzweiler, M., Benenson, R.,
  Franke, U., Roth, S., Schiele, B.: The cityscapes dataset for semantic urban
  scene understanding. In: Proceedings of the IEEE Conference on Computer
  Vision and Pattern Recognition (CVPR) (June 2016)

\bibitem{gao2019pcc}
Gao, J., Wang, Q., Li, X.: Pcc net: Perspective crowd counting via spatial
  convolutional network. IEEE Transactions on Circuits and Systems for Video
  Technology  (2019)

\bibitem{geirhos2018imagenet}
Geirhos, R., Rubisch, P., Michaelis, C., Bethge, M., Wichmann, F.A., Brendel,
  W.: Imagenet-trained cnns are biased towards texture; increasing shape bias
  improves accuracy and robustness. In: International Conference on Learning
  Representations (2018)

\bibitem{TRANCOSdataset_IbPRIA2015}
Guerrero-G{\'o}mez-Olmedo, R., Torre-Jim{\'e}nez, B., L{\'o}pez-Sastre, R.,
  Maldonado-Basc{\'o}n, S., Onoro-Rubio, D.: Extremely overlapping vehicle
  counting. In: Iberian Conference on Pattern Recognition and Image Analysis.
  pp. 423--431. Springer (2015)

\bibitem{guo2019dadnet}
Guo, D., Li, K., Zha, Z.J., Wang, M.: Dadnet: Dilated-attention-deformable
  convnet for crowd counting. In: Proceedings of the 27th ACM International
  Conference on Multimedia. pp. 1823--1832 (2019)

\bibitem{hani2020minneapple}
H{\"a}ni, N., Roy, P., Isler, V.: Minneapple: A benchmark dataset for apple
  detection and segmentation. IEEE Robotics and Automation Letters
  \textbf{5}(2),  852--858 (2020)

\bibitem{he2017mask}
He, K., Gkioxari, G., Doll{\'a}r, P., Girshick, R.: Mask r-cnn. In: Proceedings
  of the IEEE international conference on computer vision. pp. 2961--2969
  (2017)

\bibitem{he2016deep}
He, K., Zhang, X., Ren, S., Sun, J.: Deep residual learning for image
  recognition. In: Proceedings of the IEEE conference on computer vision and
  pattern recognition. pp. 770--778 (2016)

\bibitem{idrees2013multi}
Idrees, H., Saleemi, I., Seibert, C., Shah, M.: Multi-source multi-scale
  counting in extremely dense crowd images. In: Proceedings of the IEEE
  conference on computer vision and pattern recognition. pp. 2547--2554 (2013)

\bibitem{liu2016typical}
Liu, X., Song, W., Fu, L., Lv, W., Fang, Z.: Typical features of pedestrian
  spatial distribution in the inflow process. Physics Letters A
  \textbf{380}(17),  1526--1534 (2016)

\bibitem{marsden2018people}
Marsden, M., McGuinness, K., Little, S., Keogh, C.E., O'Connor, N.E.: People,
  penguins and petri dishes: Adapting object counting models to new visual
  domains and object types without forgetting. In: Proceedings of the IEEE
  Conference on Computer Vision and Pattern Recognition. pp. 8070--8079 (2018)

\bibitem{pham2015count}
Pham, V.Q., Kozakaya, T., Yamaguchi, O., Okada, R.: Count forest: Co-voting
  uncertain number of targets using random forest for crowd density estimation.
  In: Proceedings of the IEEE International Conference on Computer Vision. pp.
  3253--3261 (2015)

\bibitem{radau2009evaluation}
Radau, P., Lu, Y., Connelly, K., Paul, G., Dick, A., Wright, G.: Evaluation
  framework for algorithms segmenting short axis cardiac {MRI}. The MIDAS
  Journal-Cardiac MR Left Ventricle Segmentation Challenge  \textbf{49} (2009)

\bibitem{rahnemoonfar2017deep}
Rahnemoonfar, M., Sheppard, C.: Deep count: fruit counting based on deep
  simulated learning. Sensors  \textbf{17}(4), ~905 (2017)

\bibitem{redmon2016you}
Redmon, J., Divvala, S., Girshick, R., Farhadi, A.: You only look once:
  Unified, real-time object detection. In: Proceedings of the IEEE conference
  on computer vision and pattern recognition. pp. 779--788 (2016)

\bibitem{ren2017novel}
Ren, P., Fang, W., Djahel, S.: A novel yolo-based real-time people counting
  approach. In: 2017 international smart cities conference (ISC2). pp.~1--2.
  IEEE (2017)

\bibitem{richter2016playing}
Richter, S.R., Vineet, V., Roth, S., Koltun, V.: Playing for data: Ground truth
  from computer games. In: European conference on computer vision. pp.
  102--118. Springer (2016)

\bibitem{russakovsky2015imagenet}
Russakovsky, O., Deng, J., Su, H., Krause, J., Satheesh, S., Ma, S., Huang, Z.,
  Karpathy, A., Khosla, A., Bernstein, M., et~al.: Imagenet large scale visual
  recognition challenge. International journal of computer vision
  \textbf{115}(3),  211--252 (2015)

\bibitem{tobin2017domain}
Tobin, J., Fong, R., Ray, A., Schneider, J., Zaremba, W., Abbeel, P.: Domain
  randomization for transferring deep neural networks from simulation to the
  real world. In: 2017 IEEE/RSJ International Conference on Intelligent Robots
  and Systems (IROS). pp. 23--30. IEEE (2017)

\bibitem{wang2020DMCount}
Wang, B., Liu, H., Samaras, D., Hoai, M.: Distribution matching for crowd
  counting. In: Advances in Neural Information Processing Systems (2020)

\bibitem{wang2018deep}
Wang, M., Deng, W.: Deep visual domain adaptation: A survey. Neurocomputing
  \textbf{312},  135--153 (2018)

\bibitem{wang2019learning}
Wang, Q., Gao, J., Lin, W., Yuan, Y.: Learning from synthetic data for crowd
  counting in the wild. In: Proceedings of the IEEE conference on computer
  vision and pattern recognition. pp. 8198--8207 (2019)

\bibitem{xia2017detecting}
Xia, Y., Huang, D., Wang, Y.: Detecting smiles of young children via deep
  transfer learning. In: Proceedings of the IEEE International Conference on
  Computer Vision Workshops. pp. 1673--1681 (2017)

\bibitem{yan2019perspective}
Yan, Z., Yuan, Y., Zuo, W., Tan, X., Wang, Y., Wen, S., Ding, E.:
  Perspective-guided convolution networks for crowd counting. In: Proceedings
  of the IEEE International Conference on Computer Vision. pp. 952--961 (2019)

\bibitem{zhang2018crowd}
Zhang, L., Shi, M., Chen, Q.: Crowd counting via scale-adaptive convolutional
  neural network. In: 2018 IEEE Winter Conference on Applications of Computer
  Vision (WACV). pp. 1113--1121. IEEE (2018)

\bibitem{zhang2017fcn}
Zhang, S., Wu, G., Costeira, J.P., Moura, J.M.: {FCN}-r{LSTM}: Deep
  spatio-temporal neural networks for vehicle counting in city cameras. In:
  Proceedings of the IEEE international conference on computer vision. pp.
  3667--3676 (2017)

\bibitem{zhang2016single}
Zhang, Y., Zhou, D., Chen, S., Gao, S., Ma, Y.: Single-image crowd counting via
  multi-column convolutional neural network. In: Proceedings of the IEEE
  conference on computer vision and pattern recognition. pp. 589--597 (2016)

\bibitem{zhou2017places}
Zhou, B., Lapedriza, A., Khosla, A., Oliva, A., Torralba, A.: Places: A 10
  million image database for scene recognition. IEEE Transactions on Pattern
  Analysis and Machine Intelligence  (2017)

\end{thebibliography}
%




\end{document}